# AI and TCAD for Inverse Design and Defect Discovery: From Simple Machine Learning to LLM

Hiu Yung Wong

Electrical Engineering, San Jose State University, hiuyung.wong@sjsu.edu

***Abstract*—AI has revolutionized various engineering domains, but its impact on semiconductor device design and defect discovery is still limited, due to limited data and the curse of dimensionality. In this paper, we will discuss our work on using the Technology Computer-Aided-Design (TCAD) to generate precise data needed for machine learning (ML) to enable simulation-augmented ML. We demonstrate that with minimal domain expertise, it is possible to create a machine that performs as well as a device engineer on a specific task. We will show that auto-encoder-based machine learning models and noise engineering applied to TCAD data are effective at learning latent physics, and that the models can be seamlessly applied to experimental data. We will demonstrate how to build a device-engineer-level model step by step through various examples, including using only non-destructive electrical data to inverse-engineer the PiN diode layer thickness variations, the $Ga_2O_3$ Schottky diode doping and anode workfunction variations, and the transistor contact resistance in an inverter. Examples also include the generation of a FinFET IV/CV prediction model, the mapping between transistor images and IV curves, and the automatic calibration of TCAD parameters for a $Ga_2O_3$ Schottky diode, which can only be handled well by experienced TCAD engineers. Finally, to fully realize the potential of AI, large language models (LLMs) and multimodal LLMs (MLLMs) are believed to be necessary. We will discuss the application of LLMs to TCAD command file creation and our vision for MLLMs in automated device design and defect discovery.**

## I. Introduction

Machine learning (ML), a subfield of artificial intelligence, trains computers to make predictions by building models using data [1]. It is not possible without abundant data [2] and computing power. Since its success in early 2010, driven by the internet data boom, advances in computing power, and improvements in algorithms, it has been used across various engineering fields. The semiconductor industry is one of them. ML can be used in various areas in semiconductors. Among them, semiconductor device physics and defects have received considerable attention [3]-[7]. Compared to some other fields, device physics and defect data are scarce. There are three reasons. Firstly, the semiconductor industry is highly confidential, particularly for advanced technology nodes. There is not sufficient public data. Secondly, the total number of defective data remains low when a technology transitions from the developing phase (low volume, low yield) to the mature phase (high volume, high yield). Thirdly and most importantly, 3D tomography data is needed to build a machine that fully understands the physics and the relationship between the defect and the electrical characteristics. This is almost impossible.

The limitations can be overcome by generating data using well-calibrated simulations, such as technology computer-aided design (TCAD) and SPICE simulations. This is called TCAD-augmented ML [3] or simulation-augmented ML. In this paper, we review our simulation-augmented ML work in semiconductor inverse design, defect identification, and their applications. We will show different levels of ML and draw an analogy to human learning. Table I highlights the works.

TABLE I
Different Levels of Machine Learning and Examples used in this Paper

| Techniques Used | Examples | Corresponding Human Level |
|---|---|---|
| Simple Models (Regression, Random Forest, Neural Network) | • Inverse design of PIN Diode [3]<br>• Defect discovery in $Ga_2O_3$ Schottky barrier diode (SBD) [8] | Kindergartner |
| Noise and Principal Component Analysis + ML | • Inverse design of PIN diode [9]<br>• Defect discovery in $Ga_2O_3$ SBD[8] | Elementary Schooler |
| Simple Manifold Learning, e.g., Autoencoder | $Ga_2O_3$ SBD metal workfunction and temperature extraction [11] | Middle/High Schooler |
| Manifold Learning, e.g., Autoencoder | • Inverse design of PIN diode [10]<br>• Inverter contact resistance detection [12]<br>• FinFET surrogate model [13]<br>• Image to IV generation [14] | College student |
| Autoencoders with PINN | Automatic TCAD parameter calibration [16] | Grad. student/ Engineer |
| LMM, MLMM | TCAD code generation [25] and understanding device tomography data and physics | A team of experts |

## II. Initial Attempt of TCAD-Augmented ML

TCAD-augmented ML relies on TCAD-generated data to train a machine to discover relationships between two different sets of correlated parameters. Usually, the relationship between the electrical characteristics (such as current-voltage, IV, and capacitance-voltage, CV) and the structural parameters is of interest. In this paper, for simplicity, any parameters correlated with the electrical characteristics are referred to as structural parameters (e.g., ambient temperature).

Let us first consider a PIN diode. The IV electrical characteristics of a PIN diode are related to the thickness and doping of the p-type, intrinsic, and n-type regions of the PIN diode (structural parameters) (Fig. 1). One may use TCAD to generate various structures and simulate their IV's randomly.

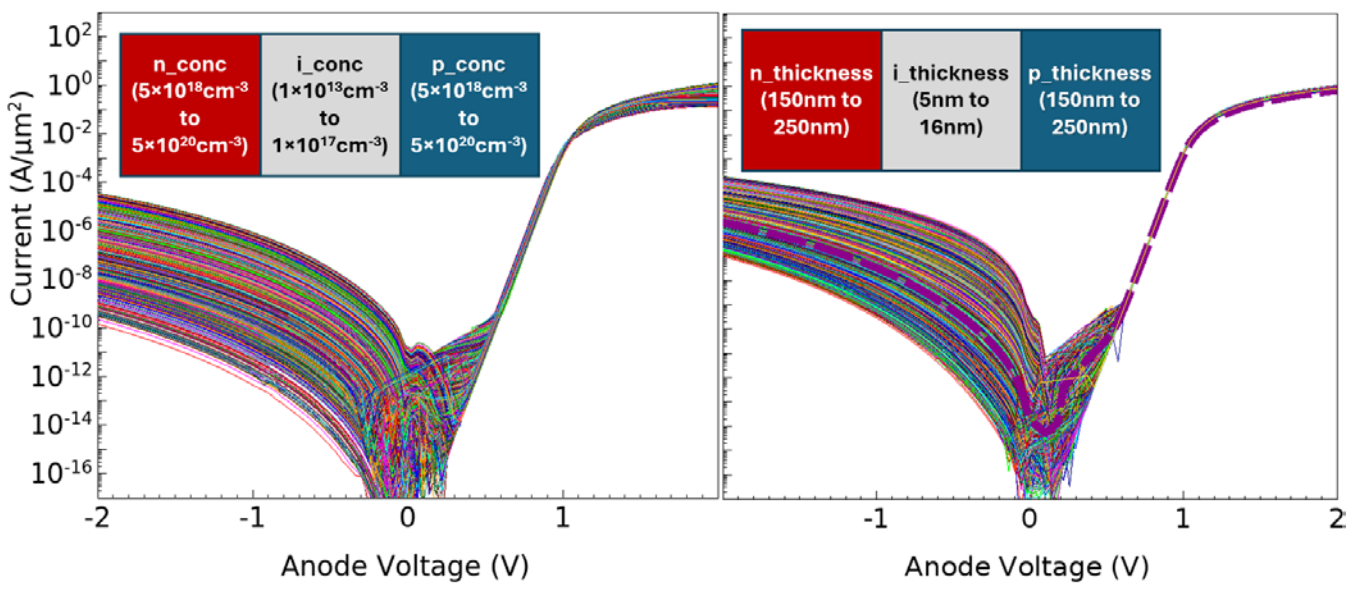

Fig. 1: IV's of 2000 devices with layer doping variations (left) and thickness variations (right) simulated using TCAD [3].

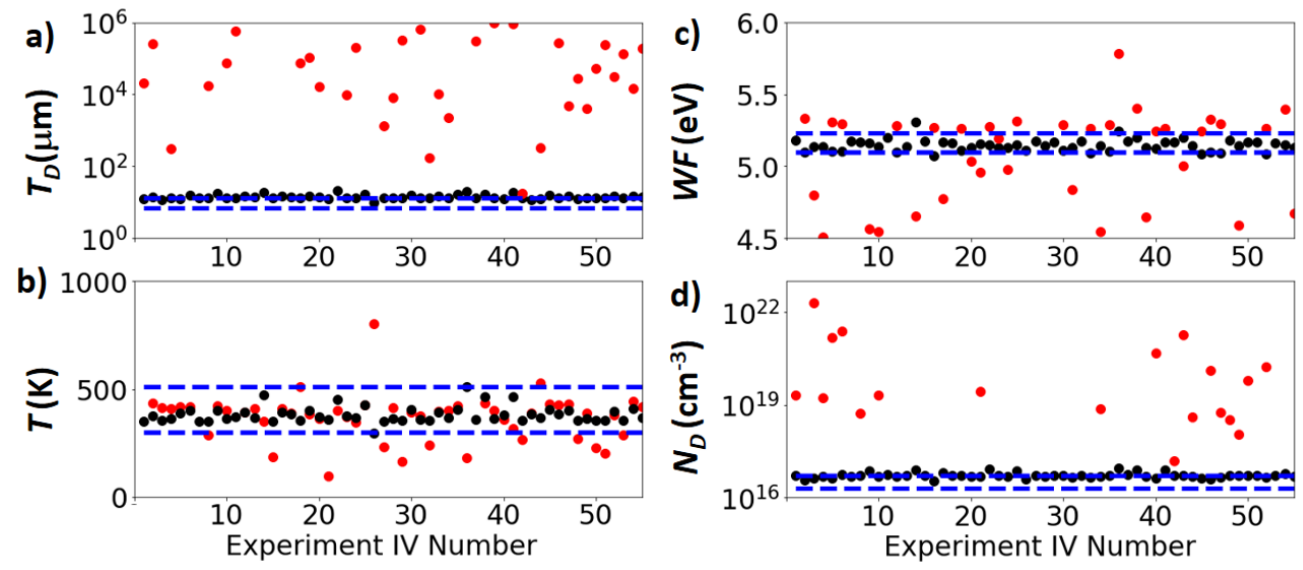

Fig. 2. Prediction of experimental SBD parameters with (black) and without (red) PCA. Blue dash lines show the expected boundaries of the parameters based on physical analysis. Some red markers (without PCA) cannot be shown because they are outside of the plotting ranges [8].

The data set is then used to train a machine to learn the correlations. Fig. 1 shows the IV's of a PIN diode as a function of doping concentration and thickness [3]. It is clear that both doping and thickness affect the reverse current. But the forward current is only affected by the doping.

Very often, the machine is used to predict the IV of unseen structures. This is dubbed as forward prediction. Sometimes, this is useful. On the other hand, many studies inaccurately claim that this approach obviates the need for long TCAD simulations and conclude that it speeds up the calculation from days to less than seconds, without taking into account that data generation requires substantial computing resources.

Another use is to deduce the thicknesses and doping levels for a given IV. This is equivalent to reverse engineering. More importantly, it is non-destructive in the experiment. If we treat the thickness and doping variations as unintentional, this provides a useful approach to discovering defects and process variations. We proposed this idea in our 2019 SISPAD paper [3]. In the same conference, Teo et al. also proposed the same concept [4]. In the study of the PIN, it is found that even a simple linear regression can perform reverse engineering very well on unseen TCAD IVs.

## III. Over-fitting and its Alleviations

The success in [3] inspired us to study the application of ML models on experimental data. In 2019, we also tried to reverse engineer the effective anode metal workfunction ($WF$), drift layer doping ($t_D$), drift layer thickness ($N_D$), and ambient temperature ($T$) of $Ga_2O_3$ Schottky barrier diodes (SBD) based on their measured IV curves using ML [8]. This was particularly useful, as $Ga_2O_3$ was an emerging material for high-power applications with significant fabrication variations. In this study, temperature-dependent TCAD parameters are well calibrated against the literature and a selected experimental curve. An ML model is then built using polynomial regression to predict the $WF$, $t_D$, $N_D$, and $T$ based on measured IV's. While it achieves excellent accuracy in predicting parameters from unseen TCAD data, it fails to predict parameters from experimental IVs, as shown in Fig. 2.

The same problem is expected to exist in the model built for the PIN diode. To verify this, besides using experimental data, one can also use simulation data generated with extra variations [9][10]. Table II shows three sets of data generated for PIN. *Set1* has only thickness variations and is used to train a machine. *Set2* and *Set3* have only doping level variations. *Set2* and *Set3* have fixed thicknesses the same as and different from the mean thicknesses of *Set1*, respectively. As demonstrated in [9] and [10], the machine trained on *Set1* predicts absurd thicknesses (e.g., negative thicknesses) when predicting the parameters using the IVs from *Set2* and *Set3*.

TABLE II: *P-I-N* DIODES SIMULATION SETUP FOR SET1, SET2 AND SET3 [9].

| Data | Thicknesses (nm) | | | Doping (cm$^{-3}$) | | |
|---|---|---|---|---|---|---|
| | n | i | p | n | i | p |
| *Set1* | 150 -250 | 5 -16 | 150 -250 | $10^{20}$ | $10^{17}$ | $10^{20}$ |
| *Set2* | 200 | 10 | 200 | $5\times10^{18}$ - $5\times10^{20}$ | $1\times10^{13}$ - $1\times10^{17}$ | $5\times10^{18}$ - $5\times10^{20}$ |
| *Set3* | 350 | 15 | 250 | $5\times10^{18}$ - $5\times10^{20}$ | $1\times10^{13}$ - $1\times10^{17}$ | $5\times10^{18}$ - $5\times10^{20}$ |

### A. Principal Components Analysis (PCA)

The machine is overfit to the TCAD data. If it were a human, we could say it is overconfident. The machine acts like a kindergartener who only memorizes knowledge without trying to understand the reasoning behind it. There is noise and hidden physics (e.g., defects) in the experimental data, which are not accounted for in the TCAD simulation. To explain what the machine has not seen in the TCAD data (e.g., nonzero current at zero bias), it "overconfidently" predicted absurd results, such as thicknesses in meters, doping concentrations larger than atomic density, and extreme temperatures. To avoid overfitting, we instructed the machine to extract four principal components from the IV data before performing regression. This is because the IVs vary as a result of changes in the four parameters. We may say that we have injected a certain domain expertise to "guide" the overconfident machine to extract features from the IVs to minimize the impact of noise and unaccounted-for physics. With that, Fig. 2 shows that all parameters are predicted to be within the expected range. Now the machine acts like an elementary schooler who can look at the data as a whole rather than memorizing every data point as an independent entity.

### B. Noise Engineering

If the machine is overconfident, another way to alleviate the issue is to add noise to the data so it understands that there is noise in the real world that it cannot explain. In [9], noise is added to the TCAD data before training. As a result, they are not perfect, and the machine is forced to extract features as memorizing the exact value of every point on the IV curve cannot be used to correlate with the parameters. Fig. 3 shows that by using simple polynomial regression, after adding noise, even without PCA analysis, the machine not only can predict the ambient temperature within a reasonable range as in Fig. 2, but also can predict the correct experimental temperature range (although the absolute values are not predicted correctly). The signal-to-noise (SNR) ratio used is 7dB. Noise engineering applies well to the PIN diode case in Table II, and is able to avoid absurd thickness predictions [9].

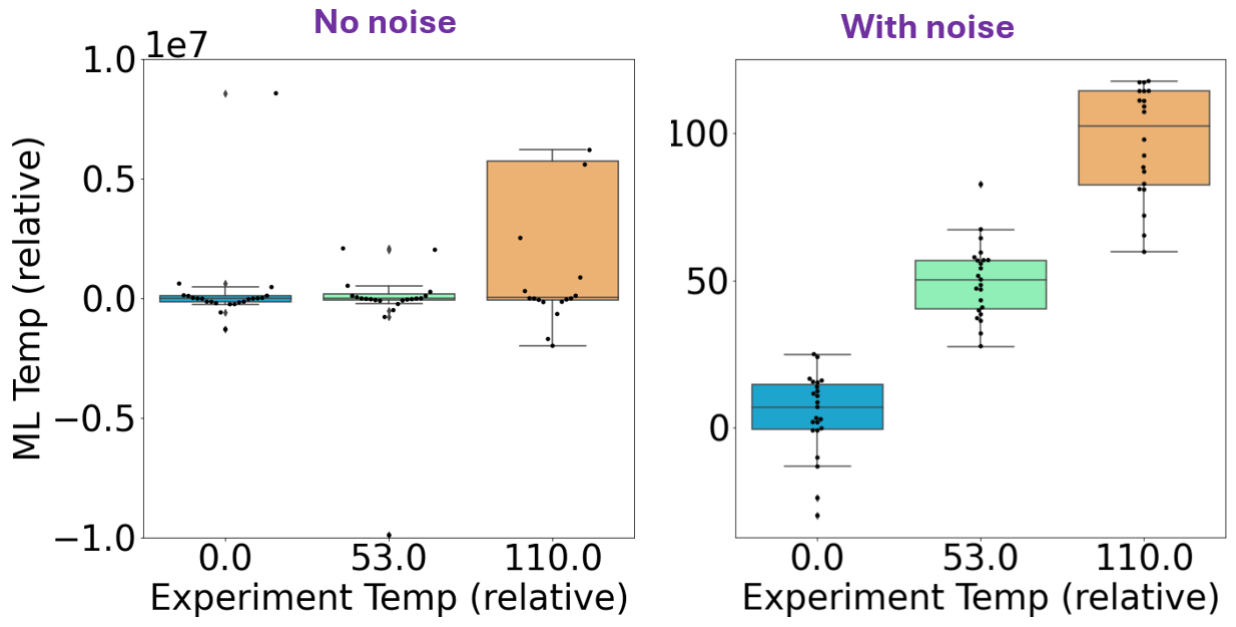

Fig. 3: Prediction of $Ga_2O_3$ SBD measurement ambient temperature using the machine trained with TCAD *IV*'s without noise (left) and with noise (right) [9].

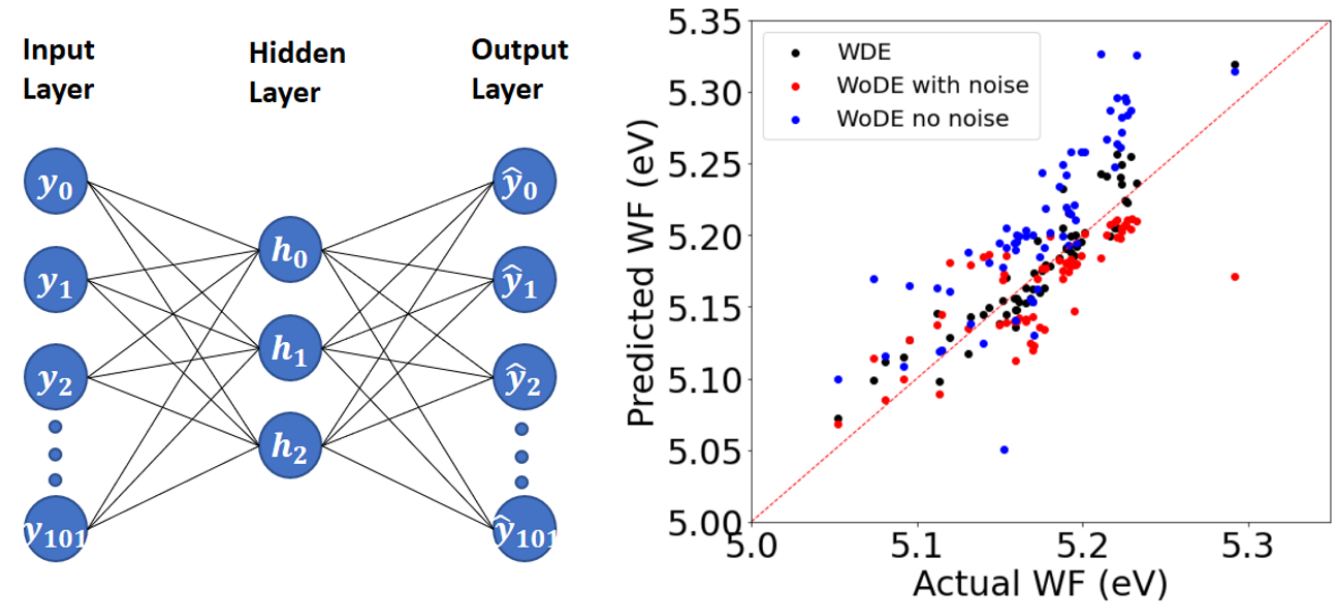

Fig. 4: Left: An autoencoder architecture with only the middle-hidden layer shown for clarity. Right: Scatter plot of $Ga_2O_3$ SBD anode metal WF prediction by the machines with (black) and without domain expertise (red and blue) [10][11].

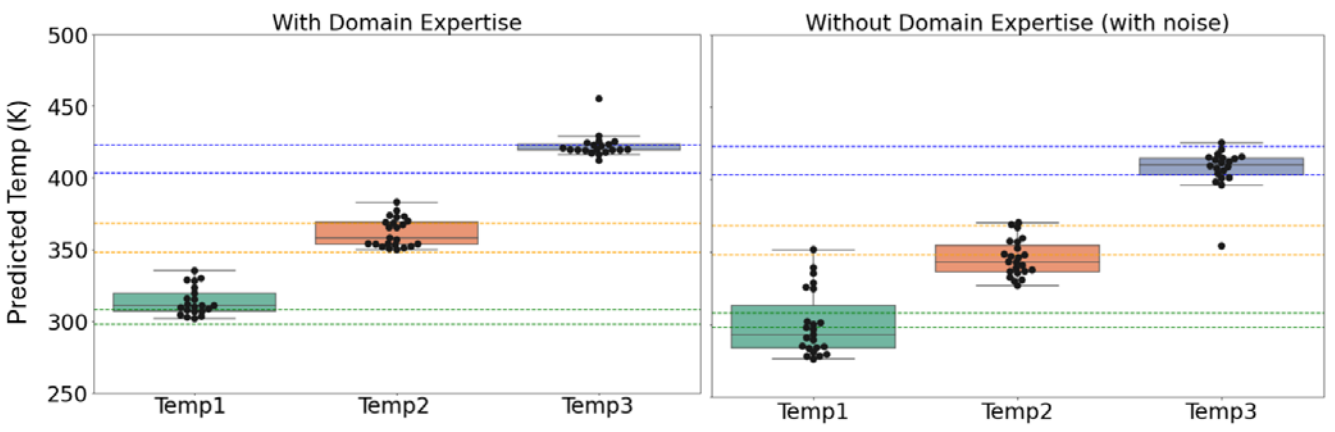

Fig. 5: Seaborn box plot of $Ga_2O_3$ SBD $T$ predictions by machines without and with domain expertise. Dotted lines are the expected temperature range of each temperature group [11].

### *C. Autoencoder*

It is natural to ask if the machine can avoid overfitting by learning the underlying physics so that it is not just "humble" but also "knowledgeable". This resembles a middle- or high schooler. However, one wants to avoid applying domain expertise in the training process. This is because applying domain expertise is time-consuming and requires experts in the field (e.g., a device physicist is needed when training a semiconductor device model). Using the $Ga_2O_3$ Schottky barrier diode as an example, if domain expertise is applied, one can extract the turn-on voltage ($V_0$) and subthreshold swing ($SS$) from the raw IVs and use them to train a machine to predict the $WF$ and $T$ of experimental IVs. The extraction scheme is based on the domain knowledge that $WF$ and $T$ directly affect $V_0$ and $SS$, respectively. This machine naturally does not have an overfitting problem because it has been "educated" to determine $WF$ and $T$ based on $V_0$ and $SS$, which are less affected by noise, instead of the IVs. Fig. 4 and Fig. 5 show that it can predict $WF$ and $T$ well with domain expertise (WDE). This resembles a teacher (expert) teaching students a skill for processing raw data to understand the correlation between electrical and structural parameters.

To obviate the need for domain expertise, one may let the machine learn the underlying physics itself (latent physics) using an autoencoder (AE). Now the machine resembles a college student. As shown in Fig. 4, an AE is just a deep neural network with the same number of nodes in the input and output layers. The goal is to train it on data so that the input and output are identical. The middle layer is constricted with a smaller number of nodes. It encodes the input data into a smaller middle latent space in the first part and can reconstruct (decode) the latent space to recover the original data. Therefore, after training, it has successfully mapped the IVs into a smaller latent space, which might have captured the underlying physics. Given that the IVs vary due to the changes in the structural parameters, one may choose the number of nodes of the middle-hidden layer to be the same as the number of structural parameters. However, the latent space usually does not have a one-to-one mapping to the structural parameters. Sometimes, more than one structural parameter can have a similar effect to the IVs. One usually sets the number of the middle-layer node to be the same as the number of structural parameters to ensure capturing the degree of freedom. Sometimes, additional nodes are used to facilitate convergence to the global minimum during training.

To extract the physical meaning of the latent space, one needs to map it to the structural parameters. Fig. 6 shows a modified AE architecture to predict $Ga_2O_3$ SBD $WF$ and $T$ based on its experimental IVs [11]. Firstly, to further reduce overfitting, noise may be applied to the TCAD training data (with or no noise in Fig. 4). Through the AE, the 52-dimensional IV curves (discretized to 52 points) are encoded into two-dimensional latent space vectors. The latent space vectors are then correlated to the parameter space vector ($WF$ and $T$) through linear regression. The experimental IV is then fed into the AE to obtain the latent space values, which are then fed into the regressor to obtain $WF$ and $T$ (green arrow in Fig. 6). Fig. 4 and Fig. 5 show that without domain expertise (WoDE) performs similarly to WDE. More importantly, the values predicted in Fig. 5 are much more accurate than those obtained by only adding noise in Fig. 3.

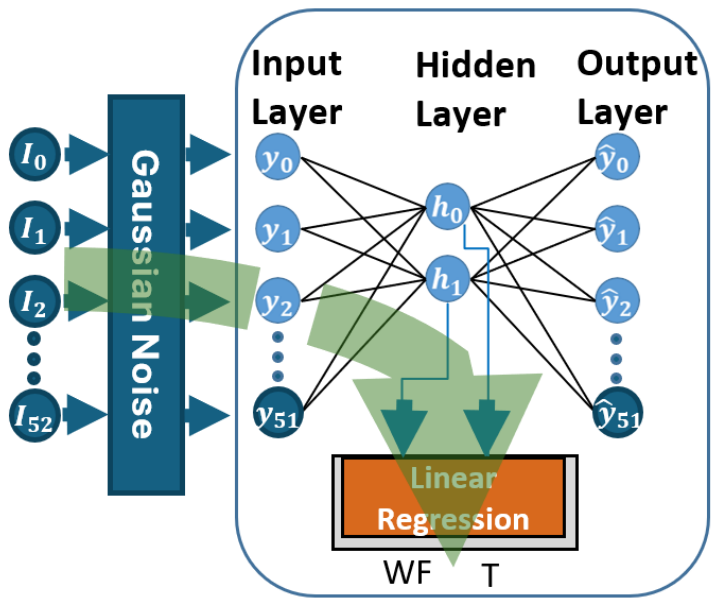

Fig. 6: ML without domain expertise. The machine is trained such that the output values equal the input values in the autoencoder and $h_0$ and $h_1$ are linearly regressed against $WF$ and $T$ [11].

## IV. AE-based Models

Autoencoder-based models enable the machine to learn the underlying physics with minimal domain expertise. This makes simulation-augmented machine learning more applicable to various domains. In this section, we will demonstrate the use of the AE-based model for defect discovery through circuit simulation, IV prediction, inverse design, and automatic calibration. It should be noted that, from an ML perspective, circuit defects, device images, and physical model parameters are the structural parameters to be correlated to electrical characteristics.

### A. Simulation-Augmented ML using SPICE

In semiconductor fabrication, test structures are placed in the scribe lines of semiconductor wafers to monitor the process. For example, one might have an inverter to monitor the overall health of the fabrication up to the local interconnect process. They can be electrically probed to make sure the transistor fabrication, contact openings, and metallization are successful. It will be ideal if the inverter voltage-transfer characteristic (VTC) curve can be used to assess the health of the fabrication process in real time. For example, if contact resistance is a major yield stopper, it would be useful to measure the VTC and rapidly deduce the contact resistance between fabrication steps. This can then be correlated to fabrication steps (e.g., contact hole critical dimensions, CD) to identify the source of the issue. In [12], we demonstrated that by training a machine on SPICE simulation data, despite the less-than-perfect SPICE model and measurement inaccuracies, one could deduce the drain contact resistances of the NMOS and PMOS in the inverter by measuring the VTC alone.

This experiment was conducted during the COVID-19 pandemic. Without access to advanced equipment and dies, instructional equipment and discrete components were used. Discrete resistors were added to mimic the contact resistances (Fig. 7). An AE-based architecture similar to Fig. 6 is used [12]. The machine has 5 hidden layers, each with 80, 50, 2, 50, and

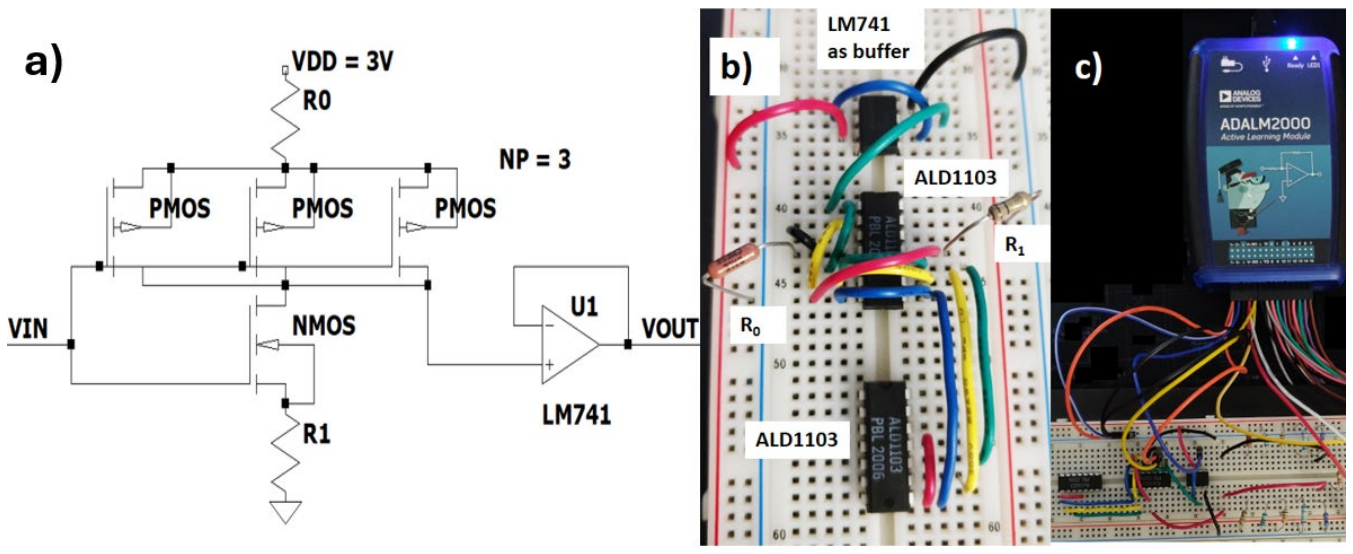

Fig. 7: a) The inverter circuit being tested. The PMOS and NMOS drain contact resistance ($R_0$ and $R_1$) are varied. Three circuits were tested with number of PMOS (NP) being 1, 2, and 3. b) The circuit on breadboard for experimental data extraction. c) The instructional grade measurement equipment used (Analog Device ADALM2000) [12].

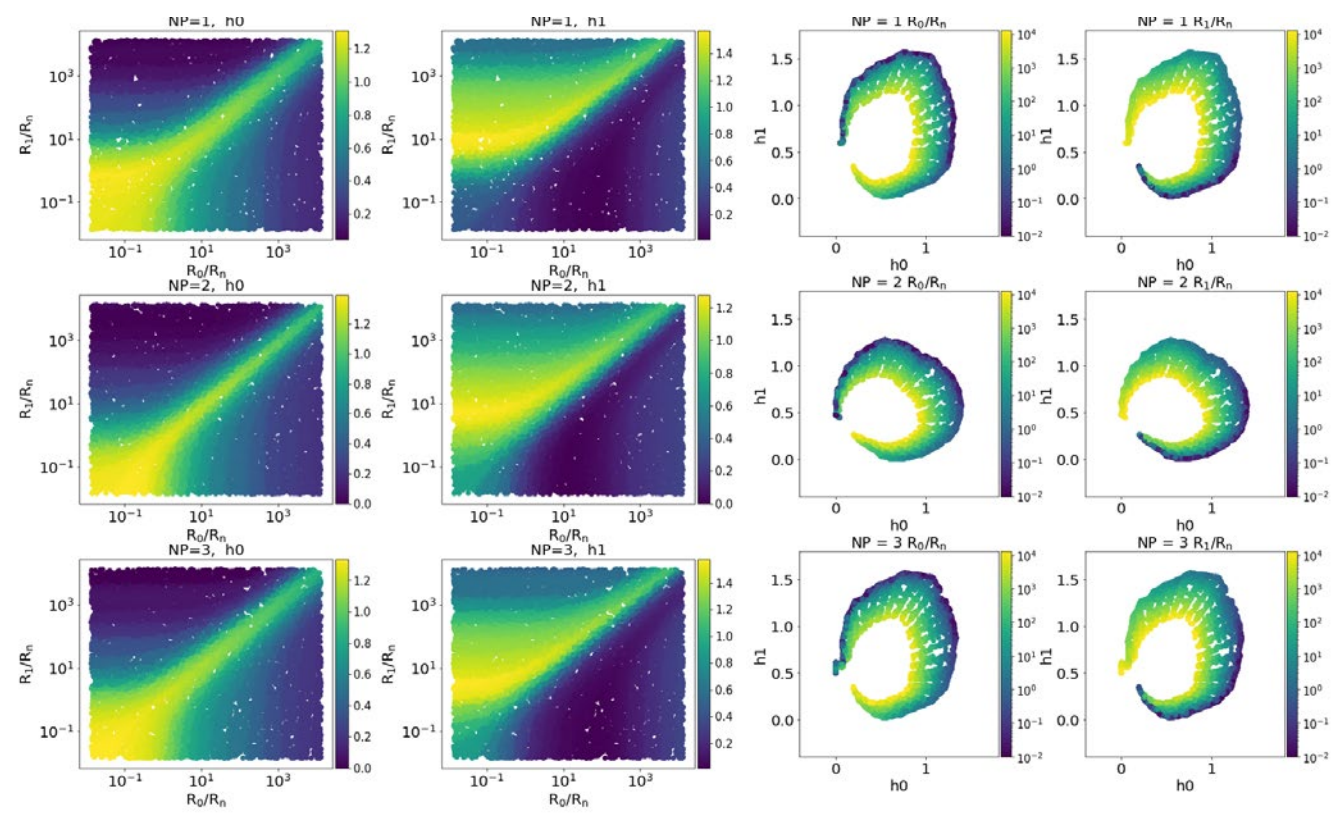

Fig. 8: $h_0$ (1st column) and $h_1$ (2nd column) as the functions of $R_0$ and $R_1$ and $R_0$ (3rd column) and $R_1$ (4th column) as the functions of $h_0$ and $h_1$ for NP = 1 (1st row), 2 (2nd row), and 3 (3rd row) [12].

80 nodes, respectively. Each VTC is discretized into 51 points and encoded as 2 latent variables, $h_0$ and $h_1$. $h_0$ and $h_1$ are correlated to $R_0$ and $R_1$ using KNN with k=3. It can predict $R_0$ and $R_1$ well (normalized to the NMOS resistance, $R_n$). More importantly, by plotting the relationship between the latent variables and the circuit parameters, one can see that the machine can capture the underlying physics well, as they have the same pattern for different circuits with varying numbers of PMOS (Fig. 8).

### B. Surrogate Model for Advanced Devices

The modified AE model can also be used to perform forward prediction by building a surrogate model. It can be used to predict the electrical characteristics for given structural parameters. Fig. 9 shows the ML model used to predict IV and CV curves for a FinFET [13]. The FinFET has 3 structural parameters being varied. The autoencoder is trained to minimize the difference between the input and output IV (CV). There are 3 nodes in the hidden layer, and they are corrected to the 3 structural parameters through third-order polynomial regression. To predict the IV or CV, the structural parameters are fed into the regression model to compute the corresponding latent values, which are then used by the decoder to predict the curves (green path in Fig. 9).

The robustness of the methodology in discovering the underlying physics is evident in the following. 1) With only 25 training data, it can train a machine (AE25) that can extract $I_{ON}$, $I_{OFF}$, transconductance ($g_m$), drain-induced-barrier-lowering (DIBL), SS, and CV with high accuracy (Tables III and IV). 2) DIBL is extracted from the curves predicted by two separate machines trained with linear and saturation IV independently. High accuracy in DIBL indicates that the machine might have learned the underlying physics. 3) A machine (AE50SD) is trained with 50 data points with a random variation of stress to mimic the unknown physics variation in the experiment. This type of data is possible to obtain from experiments during the development of an emerging technology. Tables III and IV show that high prediction accuracy can still be achieved despite the unknown variations. Therefore, such a method enables the prediction of the IV and CV curves of a new technology before its physics is well understood.

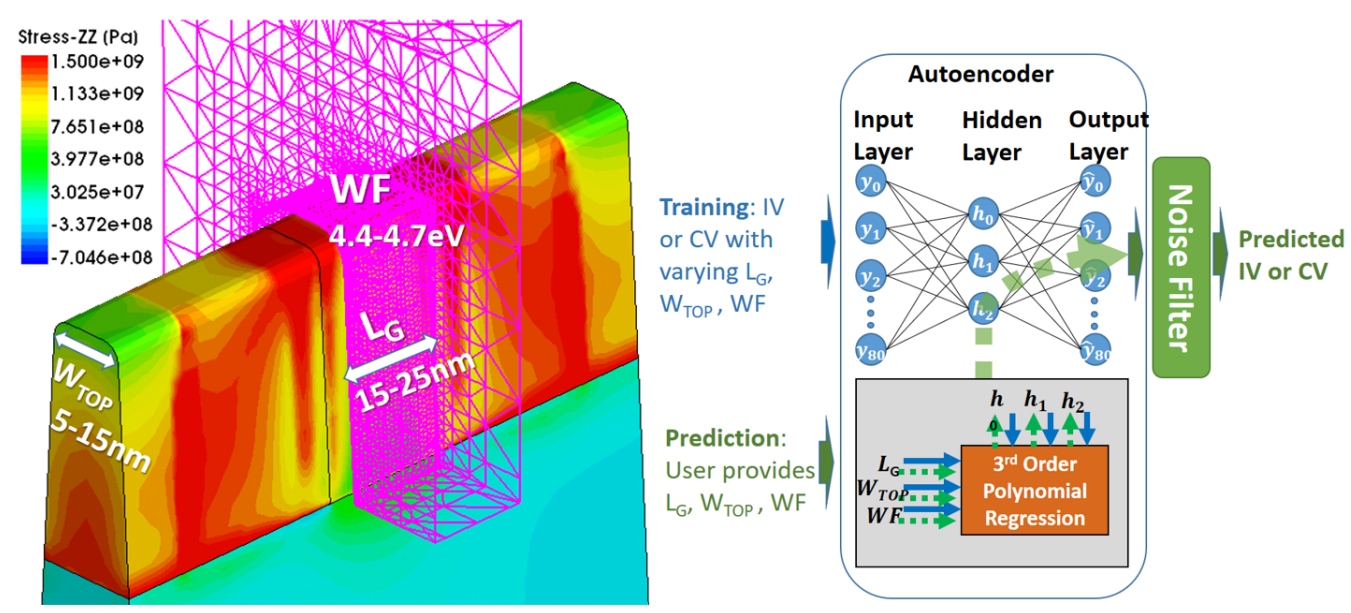

Fig. 9: Left: FinFET structure created for the simulation. Channel direction stress distribution is displayed. The 3 parameters varied and their ranges are identified. Only silicon is shown for clarity. Pink mesh is the mesh of the gate contact. Right: The ML framework used. Blue path represents the training flow during which AE and 3rd order PR are trained. The green path represents the forward prediction flow. For clarity, only a 3-layer AE is shown while a 5-layer AE is actually used [13].

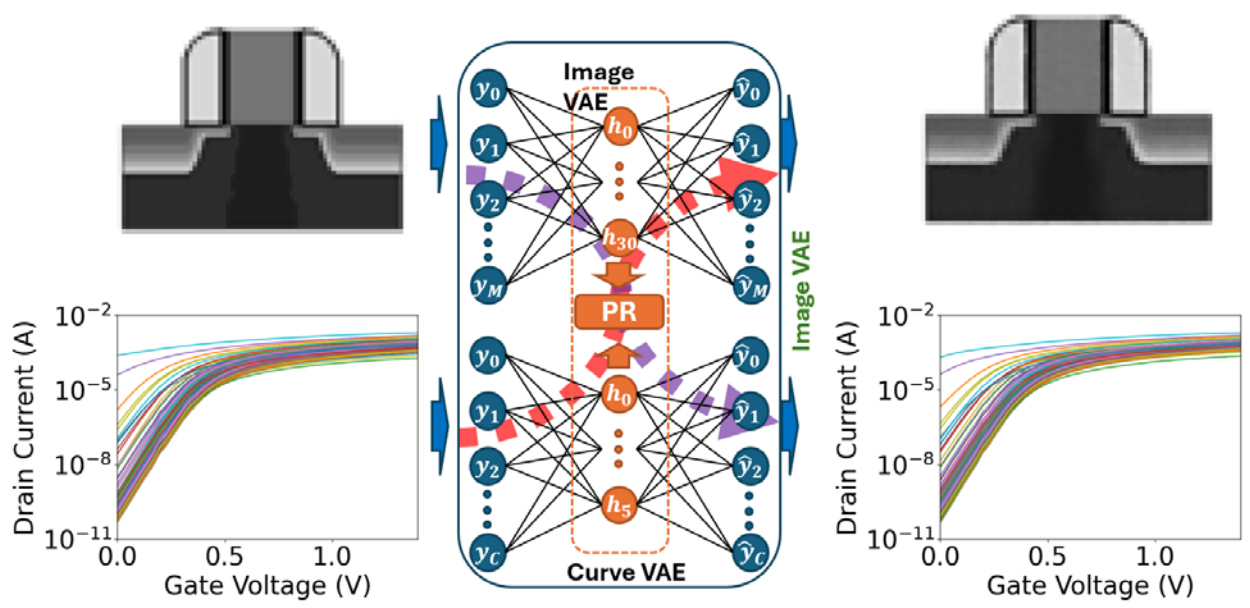

Fig. 10: A machine combining an image-VAE, a curve-VAE, and a 3rd order polynomial regressor (PR) (enclosed in the blue box) is used for image-to-curve and curve-to-image predictions. The image- (curve-) VAE is trained with the corresponding images ($I_DV_G$ curves). Only 3 layers are shown for clarity. The PR is trained to correlate the latent variables between the image-VAE and curve-VAE (enclosed in the orange box). Purple (red) thick dashed lines indicate the forward (inverse) prediction paths. One image example and a group of 100 randomly selected curves are shown as the inputs and outputs of the image- and curve-VAE, respectively [14].

TABLE III

SATURATION $I_D$ AND $G_M$ PREDICTION ACCURACY ($R^2$) OF VARIOUS MACHINES [12].

| | SATURATION $I_D$ @ $V_G$ | | | | $G_{M2}$ @$V_G$ |
|---|---|---|---|---|---|
| Machine | 0.8V | 0.4V | -0.2V | -0.6V | 0.8V |
| AE25 | 0.95 | 0.84 | 0.83 | 0.68 | 0.58 |
| AE50SD | 0.93 | 0.86 | 0.93 | 0.69 | 0.51 |

TABLE IV

OTHER METRIC PREDICTION ACCURACY ($R^2$) OF VARIOUS MACHINES [12].

| | CV Machine | | IV Machine | |
|---|---|---|---|---|
| Machine | $C_{high}$[a] | $C_{low}$[b] | DIBL | SS[c] |
| AE25 | 0.38 | 0.83 | 0.43 | 0.75 |
| AE50SD | 0.89 | 0.94 | 0.90 | 0.81 |

[a]Gate capacitance at $V_G$=0.8V, [b]Gate capacitance at $V_G$=0V, [c]Subthreshold Slope defined in the region between $10^{-9}$A to $10^{-6}$A in saturation $I_DV_G$.

### C. Correlating Images to Electrical Characteristics

To further minimize reliance on domain expertise, it is desirable to build machines that can correlate device images with their electrical characteristics, thereby obviating the need to extract structural parameters. In [14], a machine is built by combining two variational autoencoders (VAEs) and a polynomial regressor (PR) (Fig. 10). It is trained on transistor data with 5 structural parameter variations. The data contains the image and the corresponding IVs. Instead of only minimizing the reconstruction loss (difference between the input and output), a VAE is also trained by regularizing the encoding (data in the latent space) distribution. One VAE is used for IV curve reconstruction (curve VAE), and another for image reconstruction (image VAE). Therefore, convolution layers are included in the image VAE. Their latent spaces (which are expected to have captured the latent physics of the IV curves and images, respectively) are then correlated through the PR. Therefore, one can predict the IV for a given image (purple route in Fig. 10) or perform inverse design to obtain a transistor to reproduce a given IV curve (red route in Fig. 10).

Fig. 11 shows that even with hand-drawn structures (by introducing irregularity on top of a TCAD-generated structure), the machine can predict IVs that are very close to those of the corresponding TCAD-generated structure. Similarly, the machine can redraw the transistor for a given IV well [14]. It should be noted that for the same IV, there are many possible transistor designs (e.g., the same IV will be obtained for a transistor with a different poly-gate thickness). Therefore, the inverse designed transistor might not be exactly the same as the TCAD transistor used to produce the IV. The results show that it is robust to input noise and not confused by weak or irrelevant independent structural parameters.

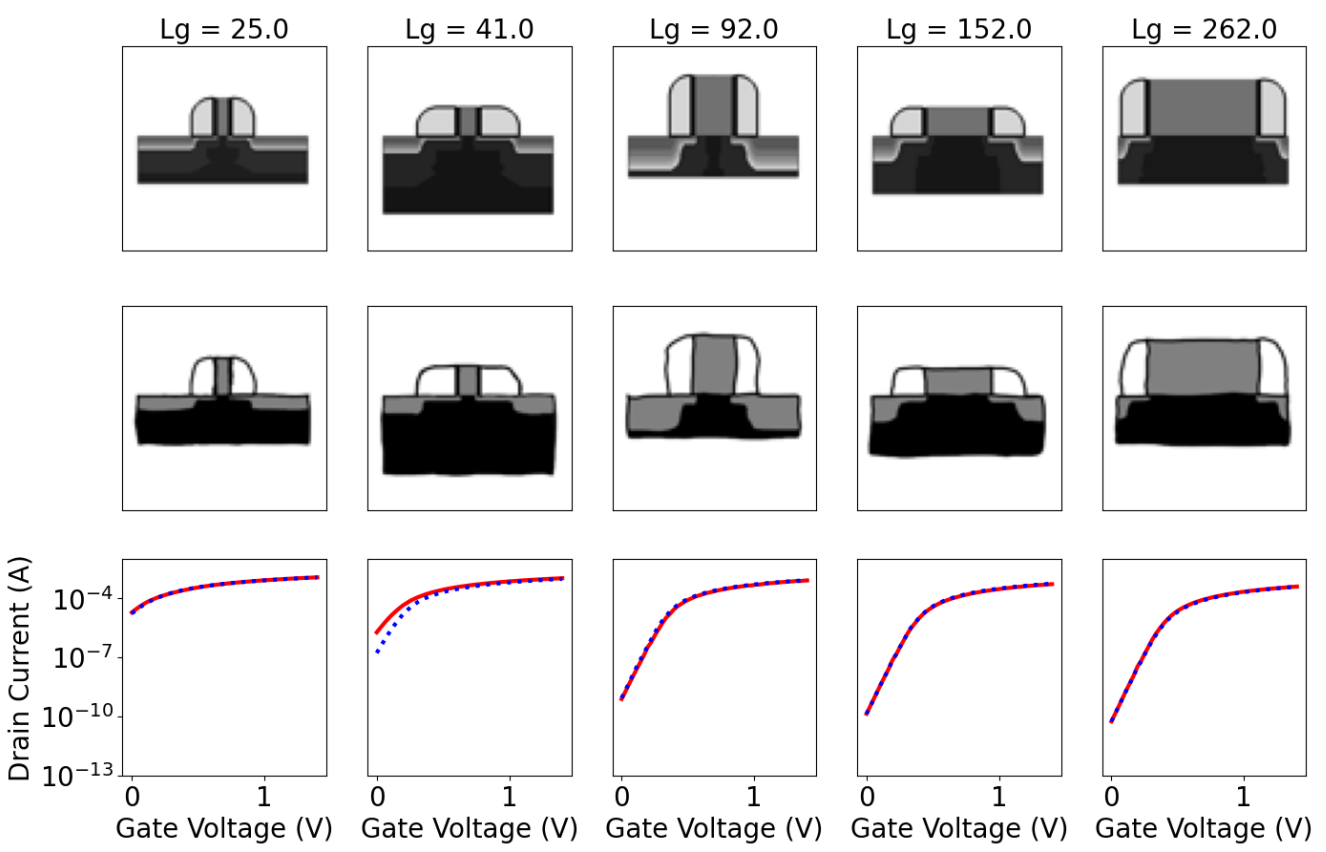

Fig. 11: Five of the structures used to test the forward prediction process. Top: TCAD-generated structures. Middle: The corresponding hand-modified structures. Bottom: TCAD simulated $I_DV_G$ based on TCAD-generated structure (dotted blue) and $I_DV_G$ from hand-drawn structure by forward design (red line). $L_G$ unit is in nm [14].

### D. Automatic TCAD Parameter Calibration and Physics-informed Neural Network (PINN)

The success of the modified-AE architecture in performing inverse design without domain expertise suggests that it may have learned the underlying physics to some extent. We therefore used it to perform autocalibration of TCAD parameters based on IVs. We first use it to automatically calibrate TCAD parameters for the Philips Unified Mobility model (PhuMob) in $Ga_2O_3$, with 6 parameters [15]. By varying the 6 PhuMob parameters, we generated 20,000 $Ga_2O_3$ SBD IVs. They are fed to train a modified AE to correlate the IVs to the parameters. This is similar to the flow in Fig. 12. However, a neural network (NN) is used to correlate the parameters to the latent space instead of a physics-informed neural network (PINN). Moreover, it does not have $WF$ variations (one fewer variable), and experimental data is not used. The latent space has a size of 10 (instead of 6) to facilitate rapid global-minimum search. Noise is added to the TCAD training data to enhance performance. Unseen TCAD IVs generated with various PhuMob parameters with noise are used to test the model. The extracted parameters are then used to run TCAD IV simulations. The resulting IVs match the expected ones well, showing a successful automatic model parameter extraction based on IV.

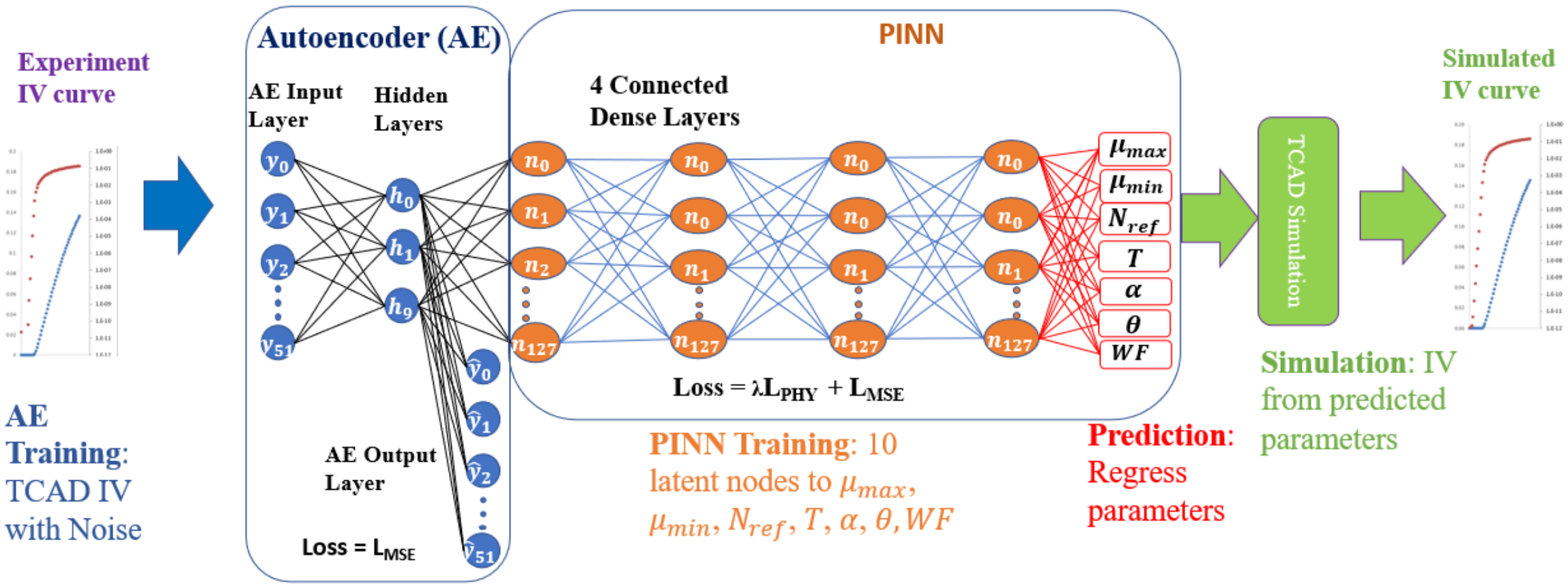

Fig. 12: Automatic TCAD parameter calibration framework. The AE and PINN are trained and tested using 5.9k simulation data. To calibrate PhuMob, 66 experimental $Ga_2O_3$ SBD IV are fed from the left through the AE and PINN to obtain the parameters (one of the curves is shown in both linear and logarithmic scales). To verify the calibration, TCAD simulations are performed to obtain the IV [16].

In [16], we increased the number of parameters to 7 and the flow is shown in Fig. 12. It was found that the machine often predicts an unrealistic set of PhuMob parameters with $\hat{\mu}_{max}$ smaller than $\hat{\mu}_{min}$. To increase the accuracy we needed to incorporate domain knowledge by including a physical loss term in the NN training. Therefore, it is equivalent to a simple PINN with physical loss $L_{\text{phy}} = \frac{1}{N}\sum_{i=1}^{N} max\big(0, \hat{\mu}_{min,i} - \hat{\mu}_{max,i}\big)$ taken into account. The machine is still trained by TCAD-generated IVs. Then, experimental IVs are fed into the machine to extract PhuMob parameters. Each experimental IV has its own $WF$ and $T$. As mentioned in Section III, they also have variations in $t_D$ and $N_D$. However, all should share the same PhuMob parameters. It is found that this methodology can successfully extract a set of PhuMob parameters that can be used in TCAD simulations to reproduce the IVs despite noise and unaccounted-for physics. Fig. 13 shows that the model with PINN (AE-PINN) performs better than the one with NN (AE-NN). More importantly, it performs similarly to what the author, who has a decade of TCAD calibration experience, has achieved after tens of hours of work. The machine now performs like a graduate student or an engineer.

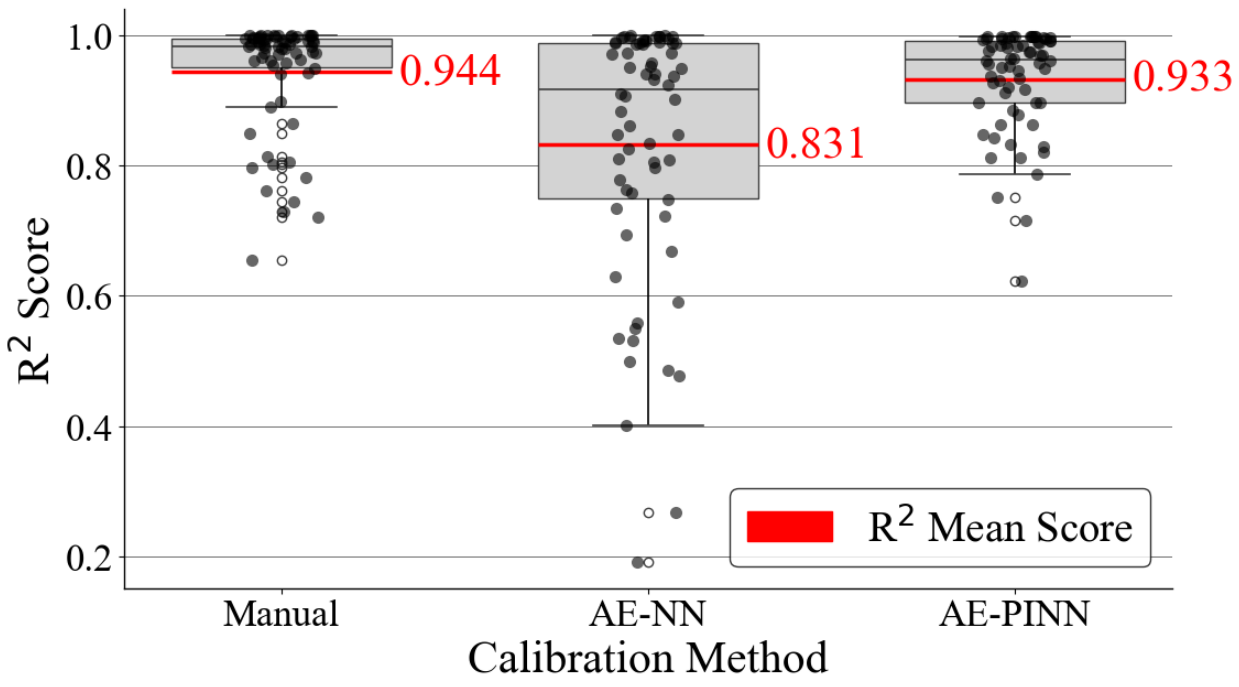

Fig. 13: Whisker box plot of the $R^2$ scores of 66 TCAD simulation I-V curves using the extracted parameters compared to the experimental I-V curves from different calibration methods - Manual, AE-NN, and AE-PINN model. The AE-PINN and Manual perform better than AE-NN. AE-PINN and Manual calibration methods are similar. Note that the manual characterization is by an expert with many years of experience in TCAD [16].

### E. Device Optimization through Surrogate Models

Surrogate models built by ML, such as the one in Section IV B, can be combined with an optimizer to optimize a device [17] [18] or even a system [19]. This is equivalent to performing an inverse design. For example, in [17], forward correlation from GaN diode guard ring parameters (Fig. 14) to electrical performance is built using an NN as a surrogate model. An NSGA-II optimizer uses the surrogate model to search for a guard ring design that will achieve the required breakdown voltage (BV) and $(V_F Q)^{-1}$ tradeoff, where $V_F$ and Q are the forward voltage and reserve capacitive charge of the diode. Fig. 14 shows the framework.

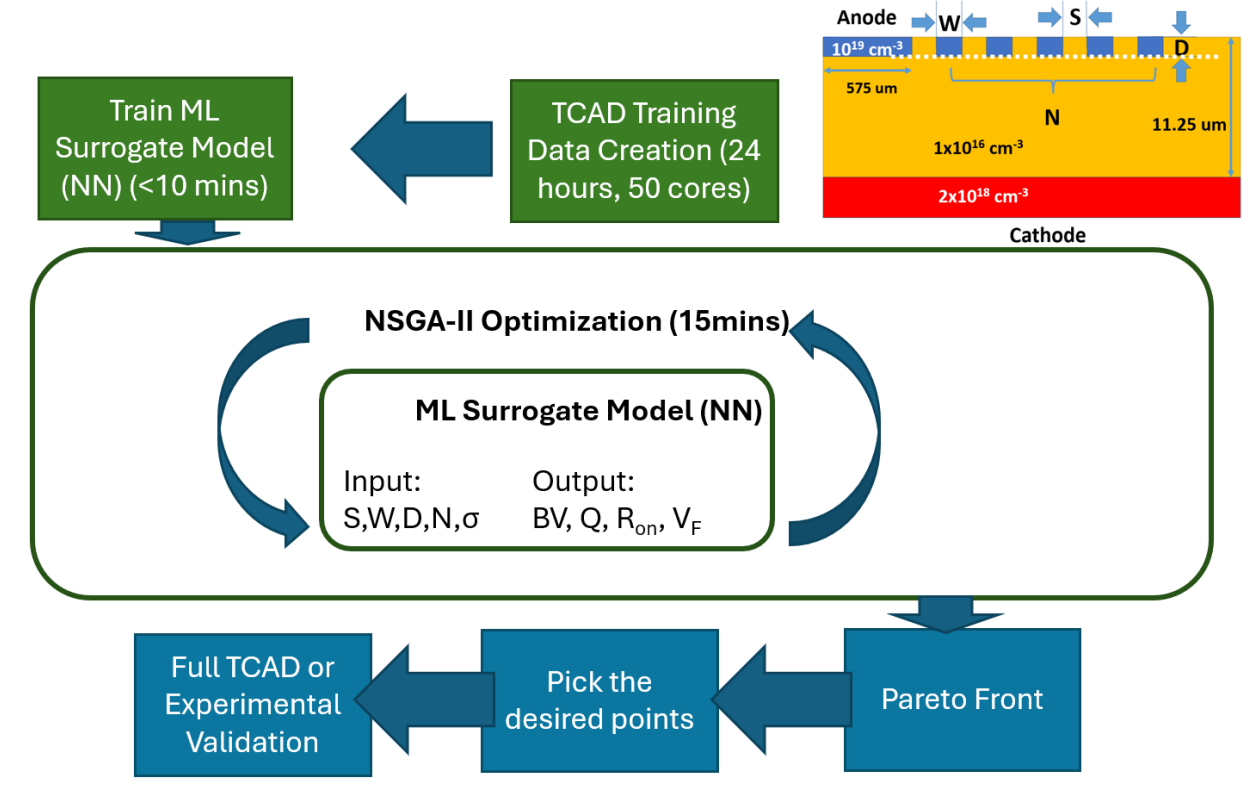

Fig. 14: The rapid inverse design framework. TCAD is used to randomly generate 266 sub-optimal structures. They are used to train a NN as the surrogate model, which will be used to find the Pareto front to discover structures with superior performance without human expertise. The time used in each step is shown [17].

## V. Limitations of Traditional ML

We demonstrated that traditional simulation-augmented machine learning, such as autoencoder-based models, can capture the underlying physics. It can be used for inverse design and defect discovery. It can also be used to build surrogate models for forward prediction and then device optimization. Besides AEs, other manifold learning techniques can also be used [20]. However, in general, it faces a few limitations. Firstly, a new model needs to be trained for every new problem. This issue is much alleviated as AE-based models require minimal domain expertise. But domain expertise is still required in some cases to avoid non-physical results [16]. Secondly, a large amount of training data needs to be generated. Intuitively, for each parameter, 3 values are needed to ensure that the machine can learn the trend. Therefore, for N parameters, $3^N$ data are needed to train a good machine. This can be alleviated by using the Latin Hypercube method [15] to reduce the amount of data required. Thirdly, some simulations are prolonged and have a low convergence (successful) rate (e.g., in BV simulations). To reduce simulation time and increase convergence rate, a simplified yet accurate strategy is needed. For example, in [18] and [19], a fast BV simulation method is used to replace the prolonged BV simulation, achieving 5 times higher throughput. But this required domain expertise and careful calibrations. Fourthly, traditional ML faces the "curse of dimensionality". So far, the largest parameter size that we have tried is only 7 [16]. The performance degrades for larger dimensions. Fifthly, the machine's ability to predict out-of-range parameters is limited, even though it has learned some of the underlying physics. For example, in [21], a generative adversarial network (GAN) has been shown to predict a parameter space with only a volume 3.7 to 4.6 times larger than the training volume. This can be improved if more domain expertise is injected, as demonstrated in [22]. Finally, many surrogate models are trained under fixed, biased conditions. For example, they can only predict $I_DV_G$ for a given $V_D$ (the discrete bias). One can generate data as a function of $V_D$ and $V_G$, but this might substantially increase the number of required data points when the number of conditions increases. To solve this problem, a conditional variational autoencoder (CVAE) can be used [23] (Fig. 15). The fixed bias (e.g., $V_D$ in $I_DV_G$ curve and $V_G$ in $I_DV_D$ curve) becomes a condition variable in the training and generation process. This allows smooth transitions between fixed biases.

## VI. Large Language Model (LLM)

Some of the aforementioned problems with traditional ML models might be alleviated by using large language models, further fine-tuned on semiconductor data. This is because the LLM has acquired some domain expertise. LLMs are also capable of reasoning and might be able to solve high-dimensional problems. An LLM has the built-in capability to handle a range of problems and provide necessary smoothing and interpolation using various data processing techniques. Therefore, an LLM has the potential to work as a strong expert team with all the necessary knowledge to solve a semiconductor problem. It is like a consulting firm, with smart learners who can solve any semiconductor problem after getting trained (fine-tuning).

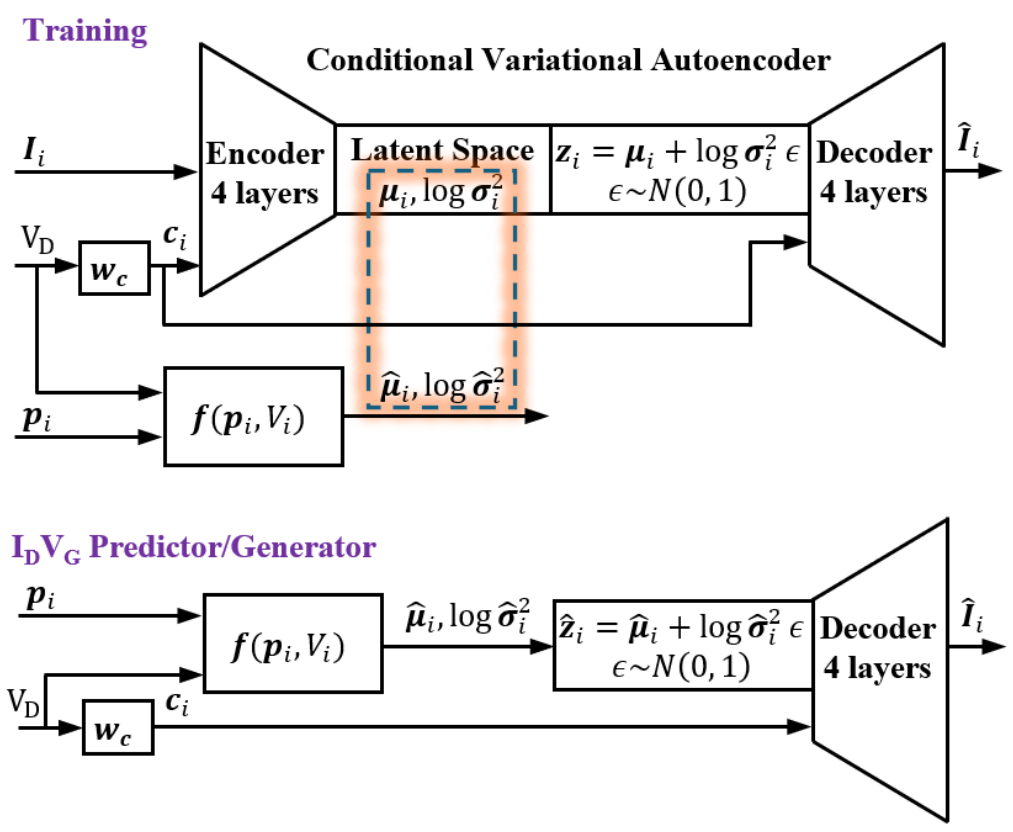

Fig. 15: Top: Training of the CVAE and the regressor $f(p_i, V_i)$ which maps the inputs to the autoencoder's latent space. Bottom: Construction of the surrogate model. $I_i$ is the $i$-th simulated $I_DV_G$ curve. $V_D$ is the condition. $z_i$ is the latent space reparametrized vector that captures the physics after training [23].

LLMs can be trained not only by text but also by other types of data, such as images. Such models are called multimodal LLMs (MLLMs). This is particularly important for semiconductors, as device physics (e.g., the distribution of the electric field and carrier density) and circuit topology affect their behavior. TCAD is again the best candidate for generating 3D tomography data, such as the spatial distribution of physical quantities that cannot be obtained from experiments. Therefore, a semiconductor database based on TCAD is needed. We have started a semiconductor database [24] but collaboration and a more coordinated effort are needed.

LLM can be used to generate input files for TCAD simulations. In [25], 7000 data points are generated to fine-tune open-source models (Llama 2 and 3) to obtain chatbots that can generate a Sentaurus Structure Editor (SSE) input file for a nanowire with 18 parameters, with English as an instruction. Each data is a pair of an English description of the nanowire structure and the corresponding SSE input file. This is likely impossible to achieve with traditional ML architectures like AEs due to the high dimensionality. But it works well with the fine-tuned LLM. It is also worth noting the history. The work was performed in early 2024 before the SISPAD deadline in April. At that point, Lamma 2, fine-tuned on the training data, outperformed ChatGPT-4, which was not fine-tuned. In-context learning (a method to aid LLMs in learning a task by providing one or more examples) also did not perform well with ChatGPT-4. After it was accepted and the final paper was being prepared in July 2024, ChatGPT4-o, with in-context learning, could already handle the question well without fine-tuning.

MLLM will likely play a critical role in future semiconductor defect discovery and reverse-engineering problem-solving. We recently performed partial fine-tuning (Low-Rank Adaptation, LoRA) and full fine-tuning on Qwen2.5-VL using the transistor structures in [14] and [24]. Transistor images are paired with transistor parameters (e.g., gate length, oxide thickness, spacer width). Note that the pictures are not annotated. The fine-tuned Qwen is able to associate the parameters with the pictures and measure the dimensions of unseen pictures.

## VII. Conclusions

We reviewed our work on the use of traditional machine learning models for semiconductor defect discovery and reverse engineering. We emphasized using TCAD and simulation data to train machines to understand the correlation between electrical characteristics and device parameters, with little to no domain expertise. By using noise engineering techniques and manifold learning (such as autoencoders), the machine can learn the underlying physics, avoid overfitting, and successfully apply to experimental data, despite the existence of noise and unaccounted physics. Finally, we discussed how LLMs and MLLMs are poised to impact the application of AI in the semiconductor field and alleviate the difficulties of traditional machine learning. However, a semiconductor database with 3D tomography is needed and can again be obtained through TCAD simulations.

## Acknowledgment

This work was supported by the National Science Foundation under Grants 2046220 and 2410694.